%% file: root.tex
\documentclass[letterpaper, 10 pt, conference]{ieeeconf}  

\IEEEoverridecommandlockouts                              

\usepackage[T1]{fontenc}

\usepackage[table]{xcolor}
\usepackage{graphics} 
\usepackage{epsfig} 
\usepackage{mathptmx} 
\usepackage{times} 
\usepackage{amsmath} 
\usepackage{amssymb}  
\usepackage{subcaption}
\usepackage{hyperref}
\usepackage{graphicx}
\usepackage{bm}
\usepackage{gensymb}
\usepackage{todonotes}
\usepackage{multirow}
\usepackage{acronym}
\usepackage{siunitx}
\usepackage{bigdelim}
\usepackage[noadjust]{cite}
\graphicspath{{Resources/Images/}}

\newcolumntype{C}[1]{>{\centering\let\newline\\\arraybackslash}m{#1}}
\DeclareMathOperator*{\argmin}{arg\,min}

\usepackage{fancyhdr}
\fancypagestyle{specialfooter}{%
  \fancyhf{}
  
  \fancyfoot[L]{\footnotesize This work was accepted by the IEEE International Conference on Robotics and Automation, Vienna, Austria, 2026.\linebreak
  © 2026 IEEE. Personal use of this material is permitted.  Permission from IEEE must be obtained for all other uses, in any current or future media, including reprinting/republishing this material for advertising or promotional purposes, creating new collective works, for resale or redistribution to servers or lists, or reuse of any copyrighted component of this work in other works.}
}

\usepackage{subfiles} 

\acrodef{gnss}[GNSS]{Global Navigation Satellite System}
\acrodef{icp}[ICP]{Iterative Closest Points}
\acrodef{imu}[IMU]{Inertial Measurement Unit}
\acrodef{ndt}[NDT]{Normal Distributions Transform}
\acrodef{rmse}[RMSE]{Root Mean Squared Error}
\acrodef{slam}[SLAM]{Simultaneous Localization and Mapping}

\DeclareMathAlphabet{\mathcal}{OMS}{cmsy}{m}{n}

\title{\LARGE \bf
CFEAR-Teach-and-Repeat: Fast and Accurate Radar-only Localization 
}

\author{Maximilian Hilger$^{1}$, Daniel Adolfsson$^{2}$, Ralf Becker$^{3}$, Henrik Andreasson$^{2}$, and Achim J. Lilienthal$^{1,2}$
\thanks{$^{1}$Chair of Perception for Intelligent Systems, Munich Institute of Robotics and Machine Intelligence, Technical University of Munich, Munich, Germany. Corresponding author: {\tt\small maximilian.hilger@tum.de}}%
\thanks{$^{2}$ Robot Navigation and Perception Lab of the AASS Research Center, Örebro University, Örebro, Sweden}%
\thanks{$^{3}$ Bosch Rexroth, Germany}%
}

\begin{document}

\maketitle
\thispagestyle{empty}
\pagestyle{empty}
\thispagestyle{specialfooter}

\subfile{Sections/abstract}

\section{INTRODUCTION}
\label{sec:introduction}
\subfile{Sections/introduction}

\section{RELATED WORK}
\label{sec:relatedwork}

\subfile{Sections/related_work}

\section{METHODOLOGY}
\label{sec:methodology}

\subfile{Sections/methodology}

\section{EVALUATION}
\label{sec:evaluation}
\subfile{Sections/evaluation}

\addtolength{\textheight}{-0.0cm}   

\section{CONCLUSION AND FUTURE WORK}
\label{sec:conclusion}
\subfile{Sections/conclusion}






\bibliographystyle{IEEEtran}
\bibliography{IEEEabrv,References.bib}

\end{document}

%% file: Sections/abstract.tex
\begin{abstract}

Reliable localization in prior maps is essential for autonomous navigation, particularly under adverse weather, where optical sensors may fail.
We present CFEAR-TR, a teach-and-repeat localization pipeline using a single spinning radar, which is designed for easily deployable, lightweight, and robust navigation in adverse conditions.
Our method localizes by jointly aligning live scans to both stored scans from the \textit{teach} mapping pass, and to a sliding window of recent live keyframes. 
This ensures accurate and robust pose estimation across different seasons and weather phenomena.
Radar scans are represented using a sparse set of oriented surface points, computed from Doppler-compensated measurements.
The map is stored in a pose graph that is traversed during localization.
Experiments on the held-out test sequences from the Boreas dataset show that CFEAR-TR can localize with an accuracy as low as 0.117 m and 0.096°, corresponding to improvements of up to 63\% over the previous state of the art, while running efficiently at 29 Hz.
These results substantially narrow the gap to lidar-level localization, particularly in heading estimation.
We make the C++ implementation of our work available to the community.

\end{abstract}

%% file: Sections/introduction.tex
In the autonomous navigation of mobile robots, accurate information about the current position is crucial.
This is of increased importance, especially in degraded settings, where GNSS reception might be blocked and sensors operating in the optical wavelength, such as cameras and lidars, fail.
Spinning radar recently emerged as a suitable sensing modality in autonomous navigation, with lidar-approaching accuracies in odometry~\cite{Adolfsson.2023cfear}, \ac{slam}~\cite{Adolfsson.2023tbv}, and autonomous path following~\cite{Qiao.2025}. 

While spinning radar has been widely studied for odometry and \ac{slam}, comparatively little attention has been given to metric localization.
In localization, the objective is to estimate the robot's pose relative to a prior map.
Current radar localization approaches achieve an accuracy of \qty{11.9}{\cm} in translation and \qty{0.27}{\degree} in rotation~\cite{Burnett.2022}.
This performance lags behind lidar, which typically localizes within \qty{5.0}{\cm} and \qty{0.04}{\degree}, with the gap especially pronounced in heading estimation~\cite{Burnett.2022}.
A common strategy to improve heading estimation, particularly in radar-based odometry, is to fuse gyroscope measurements~\cite{Burnett.2024, Li.2025, LeGentil.2025}. While this approach is effective, such multi-sensor systems pose additional requirements on extrinsic calibration, time synchronization, and bias handling. These requirements increase the setup complexity of radar-based localization in real-world deployments.        

\begin{figure}[t!]
	\centering
    \includegraphics[width = 0.47\textwidth]{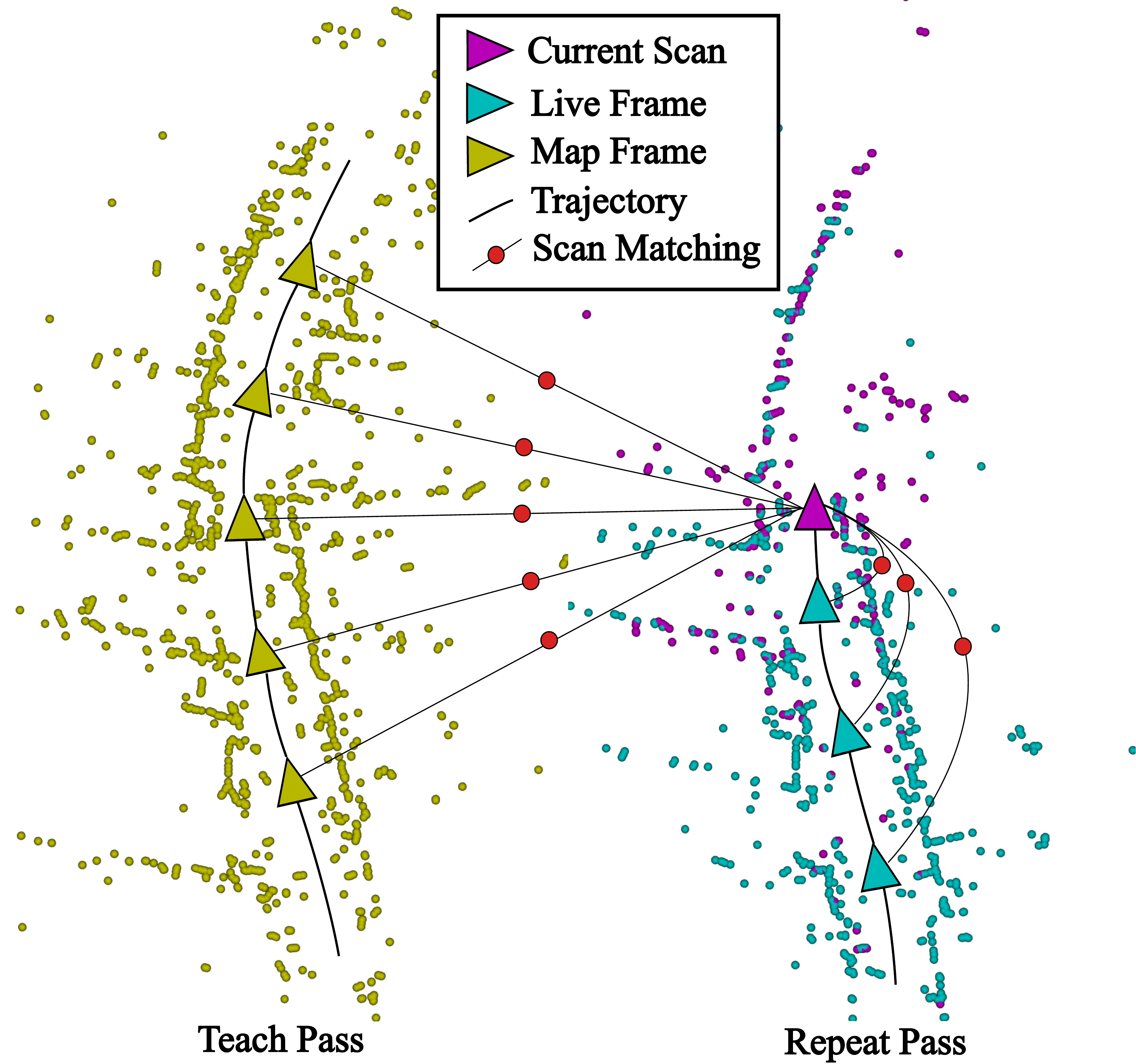}
	\caption{Visualization of the easily deployable, radar-only CFEAR-Teach-and-Repeat localization pipeline. The current scan (magenta) is simultaneously registered to keyframe scans from the teach phase (yellow) and recent live keyframes from the repeat phase (cyan). By matching sparse feature sets extracted from carefully undistorted radar scans, the system achieves both efficient and accurate localization without the need for additional sensors.
    }
	\label{fig:frontpagefigure}
    \vspace{-0mm}
\end{figure}

An alternative is to rely on radar-only localization, which offers simpler deployment and avoids the need for additional sensors.
However, existing radar-only methods remain limited. 
Prior work has applied \ac{icp}-based teach-and-repeat localization~\cite{Burnett.2022}.
Research in odometry estimation has shown that using sparse oriented surface points outperforms \ac{icp} both in terms of computational efficiency and drift~\cite{Adolfsson.2023cfear}.
Oriented surface points provide a compact and robust representation: they capture local surface geometry while supporting fast scan-to-submap registration with rich scene information. 
This makes them particularly well-suited for radar localization, yet they have not previously extended into a full localization approach.

Our approach, named CFEAR-TR (CFEAR-Teach-and-Repeat), inspired by prior work of Burnett et al.~\cite{Burnett.2022}, implements a teach-and-repeat localization framework based on the radar-only odometry approach CFEAR radar odometry~\cite{Adolfsson.2023cfear}. 
By leveraging sparse oriented surface points and a pose graph-based map structure, our system achieves accurate localization without requiring extrinsic calibration or multi-sensor fusion.
Fig.~\ref{fig:frontpagefigure} shows the pose graph structure along with its contained oriented surface points.
Our contributions are:
\begin{itemize}
    \item An efficient radar-only teach-and-repeat localization system that allows for accurate real-time route-following from sparse scan representations.
    \item A dual registration formulation combining map frames and live frames. Unlike prior work that uses odometry and localization consecutively, we optimize both against the map and recent scans simultaneously to ensure robustness to changes and reduce noise in consecutive pose estimates.
    \item An evaluation of our improved version of CFEAR radar odometry, which includes radar encoder reading, range bias, and Doppler range compensation. The presented adaptations result in a new state of the art for radar-only odometry at $0.42 \%$ translation drift, while running at 29~Hz.
    \item A systematic evaluation of parameter sensitivity and map refinement strategies. 
\end{itemize}
We demonstrate that using our approach, a localization accuracy of \qty{11.7}{\cm} and \qty{0.096}{\degree} can be achieved on hold-out test sequences on a public benchmark, cutting the previously published values in orientation by up to 63\%.
We make the C++ implementation of our work available online\footnote{\url{www.github.com/TUM-PercInS/CFEAR-TR}}.

%% file: Sections/related_work.tex
We review the related work in radar odometry and mapping (Sec.~\ref{ssec:roam}) and in radar localization (Sec.~\ref{ssec:rl}), which together frame the context of our teach-and-repeat pipeline.

\subsection{Radar Odometry and Mapping} 
\label{ssec:roam}

Spinning radar odometry incrementally estimates motion from spinning radar data.
Most approaches align live scans against local submaps to compute the motion increments.
These approaches can be divided into registration-based pipelines~\cite{Kung.2021,Adolfsson.2023cfear,Zhang.2023,Burnett.2024} and direct approaches working on image correlation~\cite{Barnes.2020mbym,Park.2020,Weston.2022,LeGentil.2025}. 
Accurate Doppler and motion compensation is a recurring requirement across methods~\cite{Burnett.2021}.
The current state-of-the-art in radar-only odometry is held by CFEAR radar odometry~\cite{Adolfsson.2023cfear}, which uses sparse oriented surface points in a scan-to-submap registration approach to estimate the sensor movement. 
We build upon this work and store the submaps in a \emph{teach} pass for usage as the reference map in the later \emph{repeat} pass.

If submaps are obtained from odometry, the poses are subject to odometry drift. 
A possible way to alleviate the drift, especially in orientation, is the use of loop closure detection.
This extends odometry pipelines to full \ac{slam}~\cite{Hong.2022,Adolfsson.2023tbv,Hilger.2024} that correct odometry for global consistency.
However, as a consequence, the local accuracy of the map might reduce -- particularly in case of incorrect loop closures, or over- or under-confident covariance estimates.
To mitigate incorrect loopclosures, Adolfsson et al.~\cite{Adolfsson.2023tbv} use an introspective loop closure detection approach that uses a combination of alignment verification, appearance similarity, and an odometry prior to reject incorrect loop closures.
We explore whether their loop closure detection applied to the pose graph in the mapping stage improves teach-and-repeat localization if multiple frames from the teach pass are used.

Another way to reduce drift, particularly in orientation, is radar-inertial fusion.
These reduce drift by integrating gyroscope measurements into odometry pipelines.
Both Burnett et al.~\cite{Burnett.2024} and Li et al.~\cite{Li.2025} use gyroscope readings to support registration-based odometry pipelines.
Le Gentil et al.~\cite{LeGentil.2025} use dense image correlation to calculate the Doppler velocity and the translation between scan and a submap.
While achieving state-of-the-art results in radar-inertial odometry, their approach cannot maintain the same accuracy without gyroscope measurements.
In general, methods using both the radar and an IMU require sensor calibration, time synchronization, and IMU bias handling, which complicates deployment.

Our work leverages CFEAR radar odometry~\cite{Adolfsson.2023cfear} as the basis for odometry and mapping. 
We enhance preprocessing by compensating for Doppler distortion, correcting range bias, and utilizing antenna angle encoder measurements. Keyframe scans derived from odometry are stored in a pose-graph structure as a map for repeat-phase localization.

\begin{figure*}[t!]
	\centering
	\includegraphics[width = 0.8\textwidth]{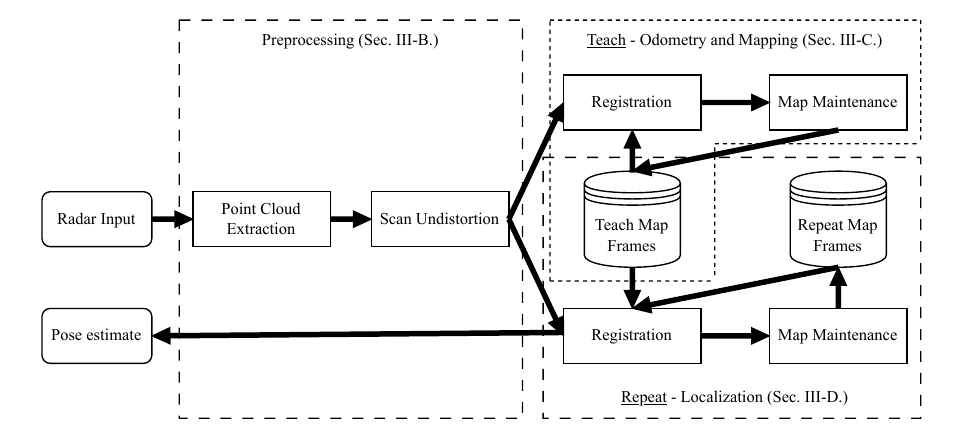}
	\caption{Architecture of our teach-and-repeat localization framework. Incoming radar scans are filtered and undistorted. In the teach pass (odometry and mapping), the scan is registered to a set of recent keyframes. In the repeat pass (localization), the scan is jointly registered with both map frames from the teach-phase and recent scans form the repeat-phase.}
	\label{fig:architecture}
\end{figure*}

\subsection{Radar Localization}
\label{ssec:rl}
Radar localization operates at different scales, including place recognition to metric localization.
Place recognition methods aim to identify previously visited areas without the need for sub-meter accuracy.
Approaches for place recognition in spinning radar encompass handcrafted descriptors~\cite{Kim.2021,Jang.2023,Gadd.2024,Kim.2024} or learning-based pipelines~\cite{Barnes.2020,Gadd.2020,Saftescu.2020,Komorowski.2021,Wang.2021,Yuan.2023}.

Metric localization, in contrast, seeks to align the live radar scan to a previously recorded map representation. 
Recent methods can be categorized into homogeneous radar-radar methods or heterogeneous radar-to-lidar approaches that make use of accurate lidar maps.
\subsubsection{Radar-to-lidar localization} 
In radar-to-lidar localization, the map is generated using lidar point clouds.
Yin et al.~\cite{Yin.2021} propose an end-to-end localization framework that directly aligns raw radar measurements to pre-built lidar maps using a differentiable measurement model.
Ma et al.~\cite{Ma.2023} use scan projection to compare radar scans to lidar scans. 
After obtaining a rough initial guess, they perform \ac{icp} to get a more accurate transformation estimate.
RaLF proposes the use of radar- and lidar feature encoders, a place recognition head, and a metric localization head to achieve global localization in a lidar map~\cite{Nayak.2024}.
All of the aforementioned methods quantify their errors relative to a global reference frame.
Burnett et al.~\cite{Burnett.2022} take a different approach to evaluate their setup: 
They implement their localization pipeline in a teach-and-repeat framework~\cite{Furgale.2010}, which evaluates the localization with regard to the closest available submap of the map.
In their work, they first compute radar odometry to create an initial guess for scan-to-map registration.
Their framework employs a motion-compensated continuous-time radar-to-lidar \ac{icp} and is able to achieve translational accuracies in the range of \qty{12}{\cm}.
In a recent follow-up, Lisus et al.~\cite{Lisus.2023} improve radar-to-lidar registration using a learnt weight mask and a differentiable \ac{icp} implementation.
While they do not employ registration in a full localization pipeline, they demonstrate improved accuracy in their scan matching.
\subsubsection{Radar-to-radar localization}
Less work exists in the area of radar-to-radar localization.
Burnett et al.'s teach-and-repeat localization~\cite{Burnett.2022} is also implemented for radar-to-radar localization using continuous-time ICP.
They achieve translational deviations in the scale of \qty{12}{\cm}, similar to radar-to-lidar localization.
Recently, Zhang et al.~\cite{Zhang.2023} proposed SDRP, where they generate a monolithic radar map based on ground truth poses. 
They propose a denosing filter and employ \ac{ndt}-based scan matching to localize the radar scans against the map.
They achieve meter-scale localization errors using their framework.

Our work presented in this paper is most similar to Burnett et al.'s radar teach-and-repeat localization~\cite{Burnett.2022}, but differs in two key aspects:
We do not use a denser set of radar detections directly, but instead, make use of a sparser set of oriented surface points  to retain efficiency. 
Additionally, we extend the formulation with a dual registration strategy that aligns both to map frames and recent live frames simultaneously, ensuring robustness under environmental change.

%% file: Sections/methodology.tex
\subsection{Radar Teach-and-Repeat Overview}
\label{ssec:componentI}

Our pipeline follows the teach-and-repeat paradigm: 
In the \emph{teach} pass, the robot traverses a previously unseen environment while generating a map.
In the \emph{repeat} pass, the robot localizes relative to this map in order to autonomously retrace the path~\cite{Furgale.2010}.
The map is stored as a pose graph, where each node contains a keyframe of radar data in the form of oriented surface points.

CFEAR-Teach-and-Repeat builds on the CFEAR radar odometry framework~\cite{Adolfsson.2023cfear}. To adapt it for teach-and-repeat localization, we make two adjustments:
First, we adapt the preprocessing by integrating Doppler compensation and using the sensor's encoder readings to convert radar images to point clouds.
While this is already done in other works~\cite{Burnett.2021, Burnett.2022, Burnett.2024}, it was not considered in CFEAR previously. 
The inclusion ensures unbiased, deskewed point clouds.
Also, we extend the odometry approach, introducing a dual registration strategy that jointly aligns live scans to (i) nearby map frames from the \emph{teach} pass and (ii) a sliding window of recent live frames, obtained during the \emph{repeat} pass. 
This aims to consistent with the taught path while preserving local consistency when the environment has changed.
Before the repeat pass, an optional global refinement step using loop closure detection and pose graph optimization can further increase global consistency of map. 
Our pipeline can be subdivided into three distinct phases: radar preprocessing (Sec.~\ref{ssec:preprocessing}), odometry and mapping (Sec.~\ref{ssec:mapping}), and localization (Sec.~\ref{ssec:localization}).
An overview is given in Fig.~\ref{fig:architecture}

\subsection{Radar Preprocessing}
\label{ssec:preprocessing}
The radar preprocessing converts raw polar scans $Z_{N_a \times N_r}$ with $N_a$ azimuth bins and $N_r$ range bins into Cartesian point clouds $\mathcal{P}$.
Its purpose is to suppress noise, correct radar-specific distortions, and produce a compact representation suitable for registration.

\subsubsection{Peak extraction}
Typical radar noise sources include multipath reflections, speckle noise, and receiver saturation. 
To suppress these, we use the \emph{k}-strongest filter~\cite{Adolfsson.2023cfear}, which, for each azimuth, selects the $k$ returns with the highest intensity exceeding a threshold $z_{min} = 60$.
Using only $k$ returns filters out multipath and receiver saturation, while the intensity threshold aims to remove the speckle noise.
The \emph{k}-strongest filter has demonstrated excellent performance in various odometry estimation pipelines~\cite{Adolfsson.2023cfear, Preston-Krebs.2025}.
The output of the filter is a set of reflections.

\subsubsection{Range correction and Cartesian conversion}
We observe two main effects, inhibiting the range readings of the spinning radar:
Doppler distortion and a static range offset.
Doppler distortion is caused by the relative velocity during chirp acquisition and is increasingly influential with higher velocities.
We compensate for the Doppler distortion using the velocity-dependent offset 
\begin{align}
    \Delta r_d(\theta_a) = \beta (v_x \cos{\theta_a} + v_y \sin{\theta_a}),
\end{align}
where the correction factor $\beta$ depends on the slope and mean of the radar's frequency ramp~\cite{Burnett.2021}, and the velocity $\begingroup \setlength\arraycolsep{1pt}\mathbf{v} = \begin{bmatrix} v_x && v_y && \omega \end{bmatrix}^T \endgroup$ is estimated from the recent history using the constant velocity assumption. In contrast to previous implementations of CFEAR~\cite{Adolfsson.2023cfear}, we do not assume equally spaced azimuth angles in the radar image, but use the encoder reading of the antenna's angle during the azimuth acquisition~$\theta_a$. 
The static range bias is corrected by an empirically tuned offset $\Delta r_r = -0.31 \si{\m}$.
The resulting polar-to-cartesian mapping can then be expressed as
\begin{align}
    \mathbf{\tilde{p}} = \begin{bmatrix} \tilde{p}_x \\ \tilde{p}_y \end{bmatrix} = (d \gamma + \Delta r_d(\theta_a) + \Delta r_r) \begin{bmatrix} \cos{\theta_a} \\ \sin{\theta_a}    \end{bmatrix}, 
\end{align}
with $d$ and $a$ describing the range and azimuth indices and $\gamma$ expressing the radar's resolution.

\subsubsection{Motion compensation}
Because the Navtech CIR radar spins at only \qty{4}{\hertz}, intra-scan motion causes distortion.
We perform motion compensation as described in CFEAR~\cite{Adolfsson.2023cfear} to remove the motion distortion, assuming constant velocity between scans.

The output of the preprocessing is a deskewed, Doppler- and range-compensated point cloud suitable for mapping and localization.
Compared to the original CFEAR~\cite{Adolfsson.2023cfear}, our preprocessing explicitly integrates Doppler compensation and uses the encoder readings of the antenna for the conversion to Cartesian space.
While these techniques are used in other radar pipelines~\cite{Burnett.2021}, these were not included in the previous evaluations of CFEAR, the impact of these improvements is discussed in Sec.~\ref{sec:evaluation}.

\subsection{Teach - Odometry and Mapping}
\label{ssec:mapping}
\begin{figure}[!t]
\centering
\begin{subfigure}[b]{0.48\textwidth}
	\includegraphics[width = 1\textwidth]{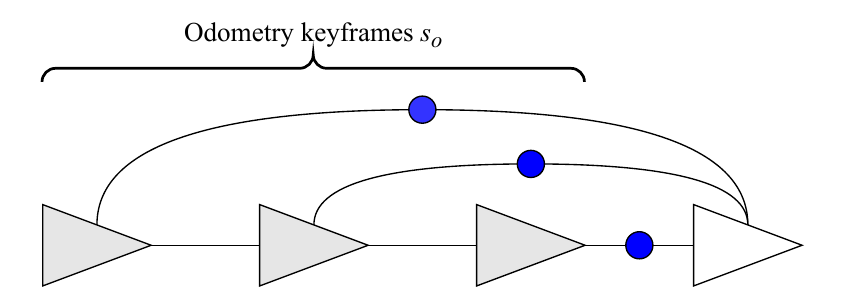}
	\caption{\emph{Teach} pass: Odometry graph using $s_o = 3$ keyframes.}
    \label{fig:odometrygraph}
   \end{subfigure}
\begin{subfigure}[b]{0.48\textwidth}
    \includegraphics[width=1\textwidth]{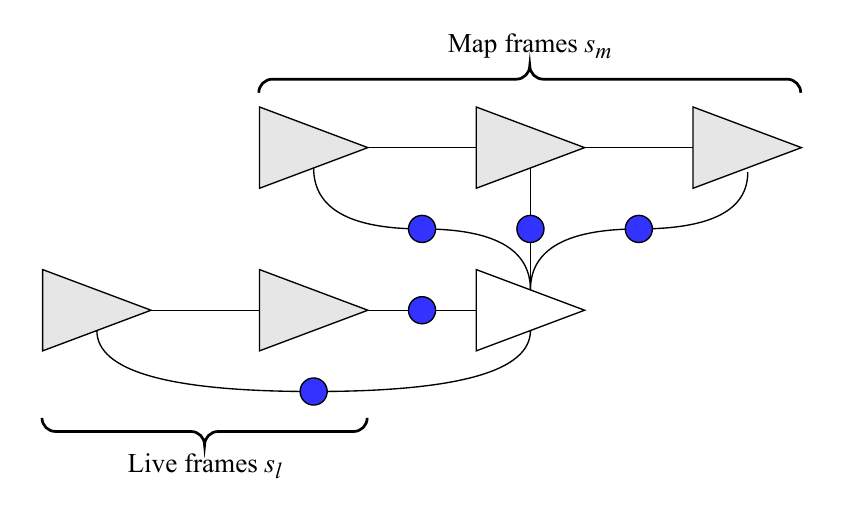}
    \caption{\emph{Repeat} pass: Localization graph using $s_m = 3$ map frames and $s_l = 2$ live frames.}
    \label{fig:localizationgraph}
\end{subfigure}
\caption{Optimization problems in (a) \emph{teach} and (b) \emph{repeat} phase, depicted as factor graphs. Grey triangles denote fixed scans; the white triangle is the optimized state. Blue dots show the scan-to-keyframe residuals.}
\label{modalities}
\end{figure}

The purpose of odometry and mapping in the \emph{teach} pass is to provide (i) incremental pose estimates for building a compact map and (ii) a set of keyframes later used in localization.
We build on CFEAR radar odometry~\cite{Adolfsson.2023cfear}, which represents scans as a sparse set of oriented surface points and performs scan-to-submap registration.
The representation $\mathcal{M}^t$ of each scan is aligned against a submap consisting of the most recent $s_o$ keyframes $\{\mathcal{M}^k\}$. 
The registration cost $f_{s2k}$ between the scan and a keyframe is given by
\begin{align}
    f_{s2k}(\mathcal{M}^k, \mathcal{M}^t, \mathbf{x}^t) =  \sum_{\forall \{i,j\} \in \mathcal{C}} w_{i,j} \mathcal{L} (g_{p2d}(m_j^k, m_i^t, \mathbf{x}^t)),
    \label{eq:robustcost}
\end{align}
where $\mathcal{C}$ denotes the set of correspondences between surface points, $w_{i,j}$ are weights scoring the similarity between surface normals, $\mathcal{L}$ the Cauchy robust loss function~\cite{Bosse.2016}, and $g_{p2d}$ the point-to-distribution residual. For implementation details, we refer the reader to~\cite{Adolfsson.2023cfear}. 
We obtain the pose estimate as 
\begin{align}
    \mathbf{x}^t = \argmin_{\mathbf{x}^t} \sum_{k=1}^{s_o} f_{s2k}(\mathcal{M}^k, \mathcal{M}^t, \mathbf{x}^t),
\end{align}
solved using the Ceres nonlinear least-squares solver~\cite{Agarwal.2022}.
The optimization problem is depicted graphically in Fig.~\ref{fig:odometrygraph}.

As described in~\cite{Adolfsson.2023cfear}, keyframes are created whenever the robot moves more than \qty{1.5}{\meter} or rotates more than \qty{5}{\degree}, ensuring sufficient geometric variation.
During the \emph{teach} pass, all keyframes are stored in a pose graph, with each node containing its pose estimate $\mathbf{x}^m$ and oriented surface points $\mathcal{M}^m$.
This pose graph forms the map used for \emph{repeat} pass localization.

As the \emph{repeat} stage uses multiple consecutive map keyframes, inaccurate transformations between keyframes due to odometry drift might inhibit localization performance.
To reduce the drift between keyframes, we, optionally, allow for optimizing the pose graph using loop closure detection.
This is done via the state-of-the-art introspective radar-only \ac{slam}, TBV SLAM~\cite{Adolfsson.2023tbv}.
We evaluate whether this drift reduction on the global scale will improve localization if using multiple map keyframes in Sec.~\ref{ssec:eval_loopclosure}.

\subsection{Repeat - Localization in Prior Map}
\label{ssec:localization}
The objective of the trajectory estimation in the \emph{repeat} pass is twofold: 
The estimated trajectory needs to (i) remain drift-free with respect to the map recorded in the \emph{teach} pass, and (ii) accurately maintain local consistency when the environment has changed (e.g., due to seasonal or dynamic effects).
To achieve this, we propose a dual registration approach.
Each incoming scan $\mathcal{M}^t$ is jointly registered against both map frames $\{\mathcal{M}^m\}$ and recent live frames $\{\mathcal{M}^l\}$.
The map frames include the closest node in the pose graph of the \emph{teach} pass and its neighbors, and anchor the trajectory to the taught path.
The live frames are a sliding window of the most recent keyframes from the \emph{repeat} pass, which preserve local consistency.
Both the number of map frames $s_m = |\{\mathcal{M}^m\}|$ and the number of live frames $s_l = |\{\mathcal{M}^l\}|$ can be tuned by the user.
An illustration of our formulation is presented in~Fig.~\ref{fig:localizationgraph}.

\begin{figure}[t!]
	\centering
	\includegraphics[width = 0.45\textwidth]{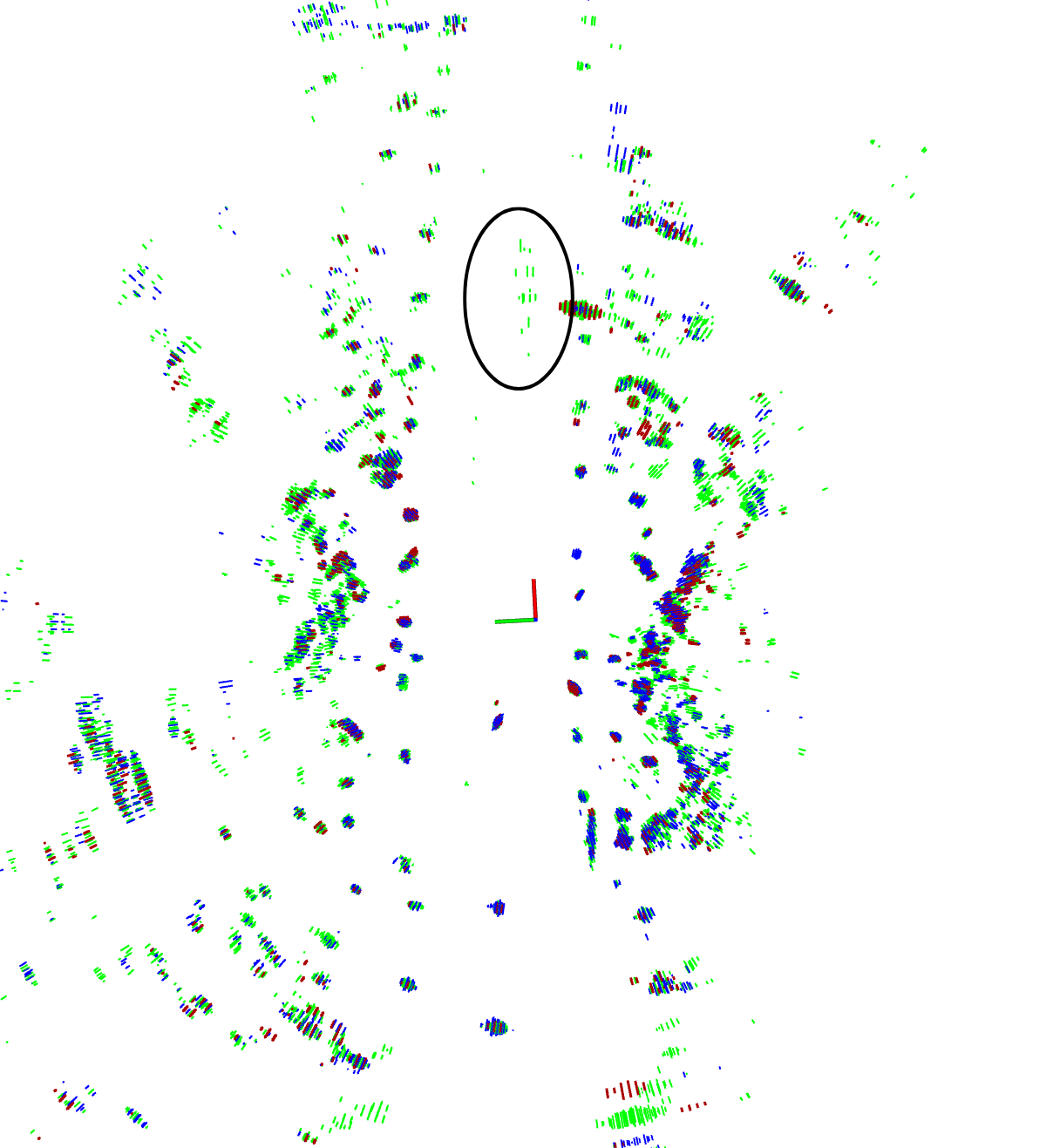}
	\caption{The live robot scan (red) is registered with map frames (green) and recent live frames (blue). Despite the presence of a dynamic object in the map (circled), the localization successfully aligns with the static environment.}
	\label{fig:pointclouds}
\end{figure}
The resulting cost function combines both terms:
\begin{align}
    \mathbf{x}^t = \argmin_{\mathbf{x}^t} \underbrace{\sum_{m=1}^{s_m} f_{s2k}(\mathcal{M}^m, \mathcal{M}^t, \mathbf{x}^t)}_{\text{scan to map frames}} + \underbrace{\sum_{l=1}^{s_l} f_{s2k}(\mathcal{M}^l, \mathcal{M}^t, \mathbf{x}^t)}_{\text{scan to live frames}},
\end{align} 
where $f_{s2k}$ denotes the robust scan-to-submap cost introduced in Eq.~\ref{eq:robustcost}.
This optimization is solved with a non-linear least squares solver.
We visualize how the live point cloud is registered with the map frames and live frames in Fig.~\ref{fig:pointclouds}.

In summary, the dual registration formulation ensures that the repeat trajectory remains consistent with the taught map while adapting locally to environmental change.

%% file: Sections/evaluation.tex
We present our experiments to show the capabilities of our method.
First, we present the experimental setup (cf.~Sec.~\ref{ssec:experimentalSetup}).
Sec.~\ref{ssec:eval_odometry} evaluates the adapted odometry pipeline used during odometry and mapping, and compares it against relevant methods.
In Sec.~\ref{ssec:eval_preprocessing}, we perform both \emph{teach} and \emph{repeat} phase experiments to investigate the importance of the presented techniques for undistorting radar scans.
After that, in Sec.~\ref{ssec:eval_parameters}, we investigate tuning of the newly introduced parameters.
We evaluate the impact of pose graph optimization on localization performance in Sec.~\ref{ssec:eval_loopclosure}.
Finally, in Sec.~\ref{ssec:eval_benchmark}, we perform a comparative evaluation against the current state of the art in radar localization.

\subsection{Experimental Setup}
\label{ssec:experimentalSetup}
We evaluate our approach on the Boreas dataset~\cite{Burnett.2023}, which was recorded in an automotive setting and contains multiple seasons, weather conditions, and lighting scenarios.
The platform is equipped with a Navtech CIR 304-H spinning radar, a 128-beam Velodyne Alpha-Prime lidar, and an Applanix POS LV GNSS-INS.
The radar has a maximum range of \qty{200}{\m} and an angular resolution of \qty{0.9}{\degree}.
During the recording, the sensor's firmware was updated.
Hence, in older sequences, the range resolution is $\gamma = 0.0596 \si{\m}$, while the range resolution of the newer sequences is a finer $\gamma = 0.04381 \si{\m}$.
Ground truth trajectories are obtained from offline fusion of RTX-\ac{gnss}, \ac{imu}, and wheel odometry, with typical accuracy of \qty{2}{\cm}-\qty{4}{\cm}.

For the teach pass, we select the sequence \texttt{2020-11-26}, which is recorded with coarser resolution.
We generate a map using radar odometry without the use of \ac{gnss} and \ac{imu} information.
Our parameter settings for the mapping sequence are the same as those used in the entry to the radar in robotics competition~\cite{Adolfsson.2024}, with the exception that we increase the number of peaks per azimuth to $k = 40$ (instead of $k = 12$).

In the repeat pass, we initialize the repeat path using the first ground truth pose.
After that, we evaluate the relative transformation from the estimated live sensor pose to the sensor pose of the closest node in the teach pass.
If not noted differently, we set the number of map frames in the registration to $s_m = 5$ and the number of live frames to $s_l = 3$.
Following the Boreas localization benchmark~\cite{Burnett.2023}, we compute the \ac{rmse} of longitudinal displacements, lateral displacements, and the heading error.
Furthermore, we keep track of the computational cost during the experiments.
All experiments have been performed on a ThinkPad P16 with 64 GB of memory and an Intel i9-13980HX (\qty{5.6}{\giga\hertz} \@ 32 cores) CPU, running a single thread.
To save computation time, we perform up to 15 experiments simultaneously.
In the following sections, we explain the different experiments we performed.

\subsection{Odometry - Comparative Evaluation}
\label{ssec:eval_odometry}
We present an evaluation of CFEAR radar odometry with modifications presented in this paper.
We follow the standard KITTI benchmark that computes drift between 100-800~m.

In the unseen test sequences of the Boreas dataset, we achieve a translation drift of $0.42\%$, and a rotation drift of $0.136 \si{\degree}/100\si{\meter}$. 
This cuts down the values reported in~\cite{Adolfsson.2024} by approximately $30\%$ in both translation and rotation. 
Notably, we even outperform most gyro-aided pipelines in both translation and rotation, scoring second in both rotation and translational drift.
Meanwhile, the odometry operates at a rate of \qty{29.4}{\hertz}, which allows execution seven times faster than the sensor rate.
This highlights the effectiveness of the improvements.
Detailed results are given in Tab.~\ref{tab:odometry}.

\input{Resources/Tables/results_odometry}

\subsection{Ablation Study - Scan Undistortion}
\label{ssec:eval_preprocessing}
In CFEAR-TR, we added a range offset, Doppler compensation, and the use of encoder measurements for the radar's azimuth (see Sec.~\ref{ssec:preprocessing}).
In this experiment, we investigate how these techniques impact odometry and localization performance.
For that, we perform an ablation study, where we compare leaving out each of the three techniques one by one. 
We compare the results to the full pipeline, and to a pipeline without all three adjustments, as described in~\cite{Adolfsson.2023cfear}.
We evaluate the teach-phase odometry on all training sequences, and store the keyframes of the mapping sequence \texttt{2020-11-26}.
The resulting odometry drift in KITTI metrics is given in Tab.~\ref{tab:odometry_ablation}.
We observe that range and Doppler offsets only moderately improve the trajectory drift.
Using the azimuth measurements also leads to just minor improvements; however, the performance gap to the full pipeline is larger than for range and Doppler offsets.
However, using all compensations together appears to have a major impact on the performance, which can be observed if comparing to the results without all adjustments.

\input{Resources/Tables/odometry_ablation}

We, furthermore, use the stored mapping sequence in the repeat phase to localize.
The resulting metrics are given in Tab.~\ref{tab:localization_ablation}.
Here, we see a different impact:
Leaving out Doppler compensation increases the longitudinal error by 165 \%.
We observe that the increase in lateral error can almost exclusively be attributed to the Doppler distortion.
Furthermore, assuming a uniform azimuth distribution leads to a three times higher heading estimation error.
The range offset has comparatively only a small influence. 
Leaving it out increases the lateral error by \qty{1}{\cm}.
This highlights the importance of proper scan undistortion for localization, which aligns with the findings explained by Burnett et al.~\cite{Burnett.2021} with regards to Doppler compensation.

\input{Resources/Tables/localization_ablation}

\subsection{Repeat-phase Registration - Parameter Study}
\label{ssec:eval_parameters}
Here, we present a parameter study of the registration component within the repeat phase.
Specifically, we investigate the number of live scans $s_l$ and the number of map frames $s_m$ used in the registration.
For that, we perform a grid search using $s_m \in \{1,3,5\}$ and $s_l \in \{1,2,3,4,5\}$.
We only test odd numbers for the number of map frames, as this ensures symmetric spacing around the closest node.
Tab.~\ref{tab:parameter_study} displays the resulting localization errors.
We observe that using a single map frame from the teach graph is not sufficient for accurate localization. 
However, adding more than three map frames does not yield better results.
This hints that with three map frames already, sufficient environmental features are available and that additional frames do not further constrain the registration.
Similarly, the number of live frames only has a modest influence on the results.
However, the number of live frames should not be larger than the number of map frames.
We can observe that in this case, errors can grow larger, which might happen due to overreliance on the drifting odometry.
All in all, our parameter study shows that already with a low number of used scans (e.g., $s_m = 3$ and $s_l = 1$), accurate overall performance is achieved.
For the remaining experiments, we select the configuration with the lowest longitudinal and heading error ($s_m = 5, s_l = 3$).
\input{Resources/Tables/parameter_study_old}

\subsection{Pose Graph Optimization in the Teach Pass}
\label{ssec:eval_loopclosure}
\begin{figure}[t!]
	\centering
	\includegraphics[width = 0.45\textwidth]{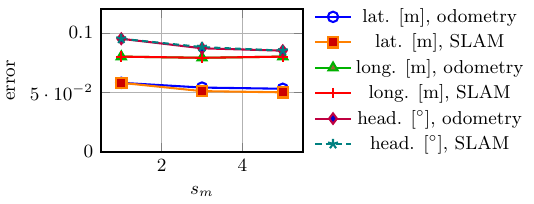}
	\caption{Localization errors in a teach graph generated using odometry and in a graph optimized with SLAM, depending on the used number of map frames $s_m$. Lateral (lat.) error is reduced, Longitudinal (long.) and Heading (head.) are not influenced.}
	\label{fig:pgo}
\end{figure}

As shown in the previous section, using multiple keyframes from the teach pass as map frames improves localization accuracy.
As these are susceptible to odometric drift, we analyze whether pose graph optimization can be used to alleviate the drift and improve localization further.
For that, we optimize the map used in the previous experiments by applying TBV SLAM~\cite{Adolfsson.2023tbv}, and localize the live radar scans within the optimized pose graph.
We evaluate our localization pipeline for varying sizes $s_m \in \{1,3,5\}$ and plot the results in Fig.~\ref{fig:pgo}, compared to the results without pose graph optimization.
We observe that the lateral localization error reduces slightly, while the longitudinal and heading errors remain at the same level.
We hypothesize that the loop closure faces the same challenge with accurate registration along the longitudinal direction.
In that case, longitudinal errors can hardly be alleviated by loop closure.
For the improved lateral accuracy, we proceed with the remaining experiments using the optimized pose graph as a reference.

\subsection{Comparative Evaluation}
\label{ssec:eval_benchmark}
\begin{figure}[t!]
	\centering
	\includegraphics[width = 0.48\textwidth]{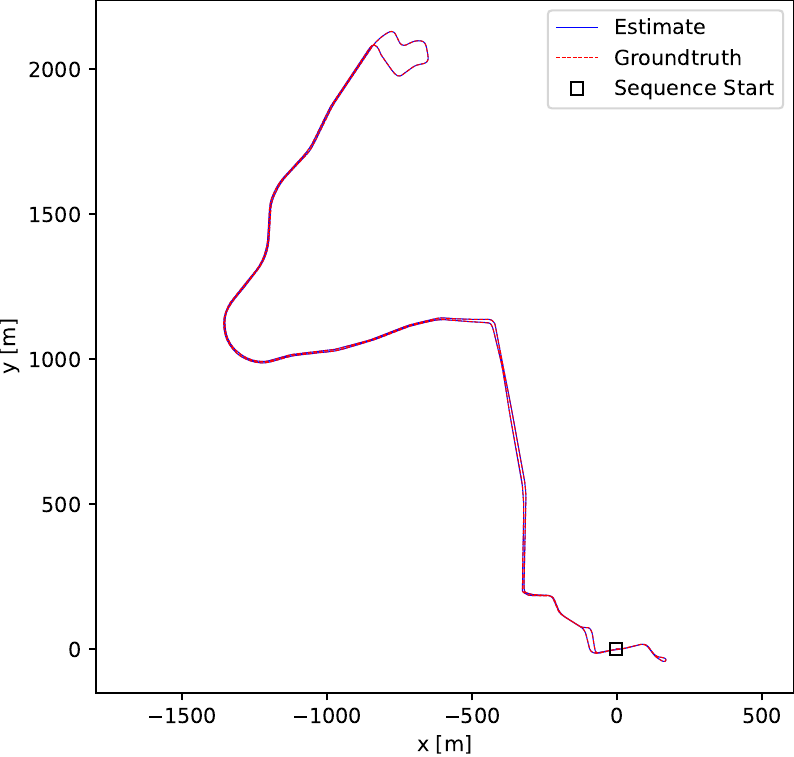}
	\caption{Trajectory estimate of CFEAR-TR on the Boreas dataset, sequence 2021-03-09.}
	\label{fig:localizationresults}
\end{figure}

\begin{figure}[t!]
	\centering
	\includegraphics[width = 0.48\textwidth]{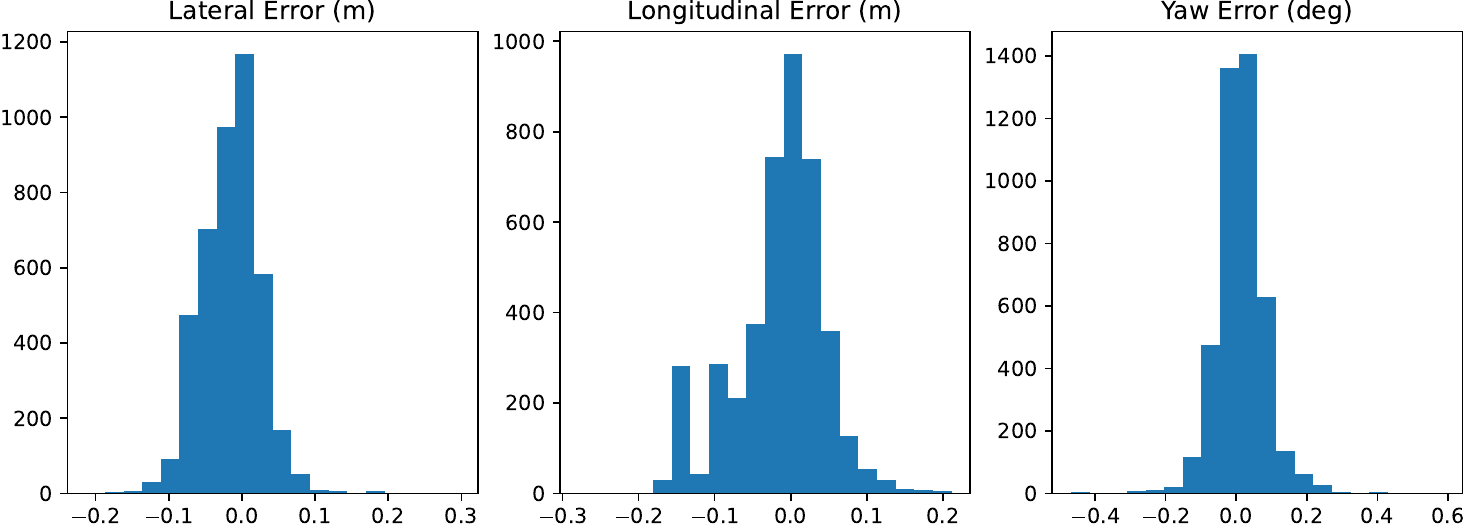}
	\caption{Distribution of localization errors using CFEAR-TR on the Boreas dataset, sequence 2021-03-09.}
	\label{fig:err_histogram}
\end{figure}

Finally, we aim to evaluate the competitiveness of our approach compared to the state of the art. 
As we showed that pose graph optimization improves the localization performance, we chose to use it as well for the comparative evaluation.
The results are shown in Tab.~\ref{tab:benchmark}, and we visualize the resulting trajectory and error distributions of an example sequence in Fig.~\ref{fig:localizationresults} and Fig.~\ref{fig:err_histogram}, respectively.

Our approach outperforms the state of the art in both lateral positioning and heading estimation. Overall localization error was measured at 0.117 m, with a lateral error as low as 0.054 m. 
We observe a substantial improvement in heading estimation, cutting down the previously reported values by $63\%$.
This is in line with the results in the odometry estimation (see Tab.~\ref{tab:odometry}).
We think that the oriented surface points, due to accumulating the points, are producing a smoother representation of the structures at the sides of the car compared to raw point clouds extracted directly from the radar scan.
The smoother representation leads to better convergence in orientation.
In longitudinal direction, we observe two small peaks in the error histogram given in Fig.~\ref{fig:err_histogram} at \qty{-0.15}{\meter} and \qty{-0.1}{\meter}.
We hypothesize that this is caused by incorrect registration results, sticking to vehicles next to the car that are moving at a similar speed.
Future work can investigate this bias in more depth.

Computation-wise, our system can operate at a rate of~\qty{29}{\hertz} on a single thread, which is many times faster than the sensor operating rate.
This demonstrates that our system can operate in real-time with low resource consumption.
Overall, due to its fast runtime and registration accuracy, we believe that CFEAR-TR forms an important step towards real-world deployment of radar-only localization.

\input{Resources/Tables/comparison_testset}

%% file: Resources/Tables/results_odometry.tex
\begin{table}[t!]
    \centering
    \caption{KITTI odometry benchmark metrics for various radar-only and gyroscope-aided approaches evaluated on test sequences from the Boreas dataset.. 
    Translational error is given in \%, rotational error in~\qty{}{\degree}/\qty{100}{\meter}. 
    Best and second best results are \textbf{bold}/\underline{underlined}.
    }
    \begin{tabular}{| l | c c c | c c |}
        \hline
        Method & radar & gyro. & accel. & transl. & rot. \\
        \hline
            VTR3 \cite{Burnett.2022}        & \checkmark & & & 1.97 & 0.565 \\
            STEAM-RO \cite{Burnett.2024}    & \checkmark & & & 1.43 & 0.410 \\
            CFEAR \cite{Adolfsson.2024}     & \checkmark & & & 0.61 & 0.206 \\
            \textbf{CFEAR-TR (ours)}        & \checkmark & & & \underline{0.42} & \underline{0.136} \\
        \hline
            CFEAR++ \cite{Li.2025}          & \checkmark & \checkmark & & 0.51 & 0.140 \\
            DRO-G \cite{LeGentil.2025}      & \checkmark & \checkmark & & \textbf{0.26} & \textbf{0.049} \\
        \hline
            STEAM-RIO \cite{Burnett.2024}   & \checkmark & \checkmark & \checkmark & 0.95 & 0.266 \\
            STEAM-RIO++ \cite{Burnett.2024} & \checkmark & \checkmark & \checkmark & 0.62 & 0.175 \\
        \hline
    \end{tabular}
    \label{tab:odometry}
\end{table}

%% file: Resources/Tables/odometry_ablation.tex
\begin{table}[t!]
	\centering
	\caption{KITTI odometry benchmark metrics on the training set, showing the effect of omitting different undistortion steps. Translational error is given in meter, and rotational error in~\qty{}{\degree}/\qty{100}{\meter}.}
	\begin{tabular}{| l | c c |}
        \hline
        & transl. error & rot. error \\
		\hline
		  full & 0.46 & 0.14 \\
        w/o doppler & 0.47 & 0.15 \\
		  w/o range offset & 0.47 & 0.15 \\
		  w/o azimuth & 0.49 & 0.16 \\
        w/o all & 0.62 & 0.19 \\
        \hline
	\end{tabular}	
	\label{tab:odometry_ablation}
\end{table}

%% file: Resources/Tables/localization_ablation.tex
\begin{table}[t!]
	\centering
	\caption{Localization errors on Boreas training sequences when different undistortion steps are omitted.}
	\begin{tabular}{| l | c c c |}
        \hline
        & lateral [m] & longitudinal [m] & heading [°] \\
		\hline
		  full & 0.053 & 0.080 & 0.085\\
        w/o doppler & 0.067 & 0.212 & 0.093 \\
		  w/o range offset & 0.063 & 0.079 & 0.088 \\
		  w/o azimuth & 0.088 & 0.087 & 0.275 \\
        w/o all & 0.099 & 0.213 & 0.277 \\
        \hline
	\end{tabular}	
	\label{tab:localization_ablation}
\end{table}

%% file: Resources/Tables/parameter_study_old.tex
\begin{table}[t!]
	\centering
	\caption{Localization errors for different amounts of map frames $s_m$ and live frames $s_l$. The RSME is given as (lateral~[\qty{}{\meter}]~/~longitudinal~[\qty{}{\meter}]~/~heading~[\qty{}{\degree}]). Best and second best results are \textbf{bold}/\underline{underlined}.}
	\begin{tabular}{| l | c  | c | c |}
        \hline
        & $s_m = 1$ & $s_m = 3$ & $s_m =5$ \\
		\hline
		$s_l = 1$ & 0.057/0.109/0.110 & 0.055/0.080/0.089 & \underline{0.054}/0.081/\underline{0.087} \\
		$s_l = 2$ & 0.065/0.081/0.105 & \underline{0.054}/\underline{0.079}/\underline{0.087} & \underline{0.054}/0.081/\underline{0.087} \\
		$s_l = 3$ & 0.058/0.080/0.095 & \underline{0.054}/\underline{0.079}/\underline{0.087} & \textbf{0.053}/0.080/\textbf{0.085} \\
		$s_l = 4$ & 0.072/0.081/0.105 & 0.067/0.080/0.098 & \underline{0.054}/\underline{0.079}/\textbf{0.085} \\
		$s_l = 5$ & 0.081/0.082/0.108 & \underline{0.054}/\underline{0.079}/\textbf{0.085} & \underline{0.054}/\textbf{0.078}/\underline{0.087} \\
        \hline
	\end{tabular}	
	\label{tab:parameter_study}
\end{table}

%% file: Resources/Tables/comparison_testset.tex
\begin{table}[t!]
	\centering
	\caption{Comparison with the state of the art approach VTR3~\cite{Burnett.2022}. The RMSE is given as (lateral~[\qty{}{\meter}]~/~longitudinal~[\qty{}{\meter}]~/~heading~[\qty{}{\degree}]).}
	\begin{tabular}{| l | c  | c | c | c | c | c |}
        \hline
        & \multicolumn{3}{c|}{\textbf{CFEAR-TR} (ours)} & \multicolumn{3}{c|}{VTR3 \cite{Burnett.2022}}\\
        & lat. & long. & head. & lat. & long. & head. \\
		\hline
		2020-12-04 & 0.070 & 0.088 & 0.088 & 0.072 & 0.082 & 0.211 \\
        2021-01-26 & 0.044 & 0.050 & 0.102 & 0.048 & 0.055 & 0.227 \\
		2021-02-09 & 0.063 & 0.056 & 0.105 & 0.053 & 0.051 & 0.235 \\
		2021-03-09 & 0.043 & 0.061 & 0.072 & 0.051 & 0.053 & 0.233 \\
		2021-06-29 & 0.052 & 0.094 & 0.083 & 0.069 & 0.095 & 0.246 \\
		2021-09-08 & 0.056 & 0.107 & 0.119 & 0.067 & 0.110 & 0.269 \\
		2021-10-05 & 0.052 & 0.119 & 0.093 & 0.069 & 0.109 & 0.288 \\
		2021-10-26 & 0.053 & 0.124 & 0.095 & 0.060 & 0.119 & 0.283 \\
		2021-11-06 & 0.055 & 0.157 & 0.093 & 0.062 & 0.155 & 0.256 \\
		2021-11-28 & 0.049 & 0.182 & 0.109 & 0.058 & 0.190 & 0.436 \\
        \hline
        \textbf{mean} & \textbf{0.054} & \textbf{0.104} & \textbf{0.096}& \textbf{0.061} & \textbf{0.102} & \textbf{0.268} \\
        \hline
        exec. time [\si{\second}] & \multicolumn{3}{c|}{0.034} & \multicolumn{3}{c|}{0.075} \\
        \hline
	\end{tabular}	
	\label{tab:benchmark}
\end{table}

%% file: Sections/conclusion.tex
In this work, we presented CFEAR-TR, a fast and efficient radar-only teach-and-repeat localization system that builds on CFEAR radar odometry.
Compared to the original formulation, we enhanced preprocessing with Doppler compensation and encoder-based azimuth readings, yielding more consistent detection sets.
Our localization further employs a dual registration strategy that jointly registers the current scan to nearby map frames from the teach pass and to a sliding window of recent live keyframes.
Using pose graph optimization as a global refinement further increases the consistency of the map, as demonstrated by our experiments.
Evaluation on the Boreas dataset indicates that the system achieves a localization accuracy as low as \qty{0.117} {\meter} and \qty{0.096}{\degree}, corresponding to an improvement of 63\% in heading estimation over the previous state of the art in radar-only localization. 
These results substantially narrow the gap to lidar-level localization. Future work aims to investigate whether Doppler-enabled spinning radar can be used to close the performance gap to lidar in the longitudinal direction, given its resilience in geometrically underconstrained environments. In addition, our method is computationally efficient, making it well-suited for radars with higher spin and measurement rates, which could further help close the gap.